\documentclass[sigconf]{acmart}

\copyrightyear{2024}
\acmYear{2024}
\setcopyright{rightsretained}
\acmConference[SIGMOD-Companion '24]{Companion of the 2024 International Conference on Management of Data}{June 9--15, 2024}{Santiago, AA, Chile}
\acmBooktitle{Companion of the 2024 International Conference on Management of Data (SIGMOD-Companion '24), June 9--15, 2024, Santiago, AA, Chile}
\acmDOI{10.1145/3663741.3664785}
\acmISBN{979-8-4007-0422-2/24/06}

\copyrightyear{2024}
\acmYear{2024}
\setcopyright{acmlicensed}\acmConference[BiDEDE '24]{International Workshop on Big Data in Emergent Distributed Environments }{June 9--15, 2024}{Santiago, AA, Chile}
\acmBooktitle{International Workshop on Big Data in Emergent Distributed Environments (BiDEDE '24), June 9--15, 2024, Santiago, AA, Chile}
\acmDOI{10.1145/3663741.3664785}
\acmISBN{979-8-4007-0679-0/24/06}

\begin{document}

\title{Are Large Language Models the New Interface for Data Pipelines?}

\author{Sylvio Barbon Junior}
\email{sylvio.barbonjunior@units.it}
\orcid{0000-0002-4988-0702}
\authornotemark[1]
\affiliation{%
  \institution{University of Trieste}
  \country{Italy}
}

\author{Paolo Ceravolo}
\email{paolo.ceravolo@unimi.it}
\affiliation{%
  \institution{University of Milan}
  \country{Italy}
}

\author{Sven Groppe}
\email{sven.groppe@uni-luebeck.de}
\affiliation{%
  \institution{University of Lübeck}
  \country{Germany}
}

\author{Mustafa Jarrar}
\email{mjarrar@birzeit.edu}
\affiliation{%
  \institution{Birzeit University}
  \country{Palestine}
}

\author{Samira Maghool}
\email{samira.maghool@unimi.it}
\affiliation{%
  \institution{University of Milan}
  \country{Italy}
}

\author{Florence Sèdes}
\email{sedes@irit.fr}
\affiliation{%
  \institution{IRIT, University Toulouse 3}
  \country{France}
}

\author{Soror Sahri}
\email{soror.sahri@u-paris.fr}
\affiliation{%
  \institution{Université Paris Cité}
  \country{France}
}

\author{Maurice van Keulen}
\email{m.vankeulen@utwente.nl}
\affiliation{%
  \institution{University of Twente}
  \country{The Netherlands}
}

\renewcommand{\shortauthors}{Barbon Junior, et al.}

\begin{abstract}
A Language Model is a term that encompasses various types of models designed to understand and generate human communication. Large Language Models (LLMs) have gained significant attention due to their ability to process text with human-like fluency and coherence, making them valuable for a wide range of data-related tasks fashioned as pipelines. The capabilities of LLMs in natural language understanding and generation, combined with their scalability, versatility, and state-of-the-art performance, enable innovative applications across various AI-related fields, including eXplainable Artificial Intelligence (XAI), Automated Machine Learning (AutoML), and Knowledge Graphs (KG). Furthermore, we believe these models can extract valuable insights and make data-driven decisions at scale, a practice commonly referred to as Big Data Analytics (BDA). In this position paper, we provide some discussions in the direction of unlocking synergies among these technologies, which can lead to more powerful and intelligent AI solutions, driving improvements in data pipelines across a wide range of applications and domains integrating humans, computers, and knowledge.

\end{abstract}



\received{20 February 2024}  
\received[revised]{12 March 2024} 
\received[accepted]{5 June 2024}  

\maketitle

\section{Introduction}
\begin{figure}[ht!]
    \centering
    \includegraphics[width=0.4\textwidth]{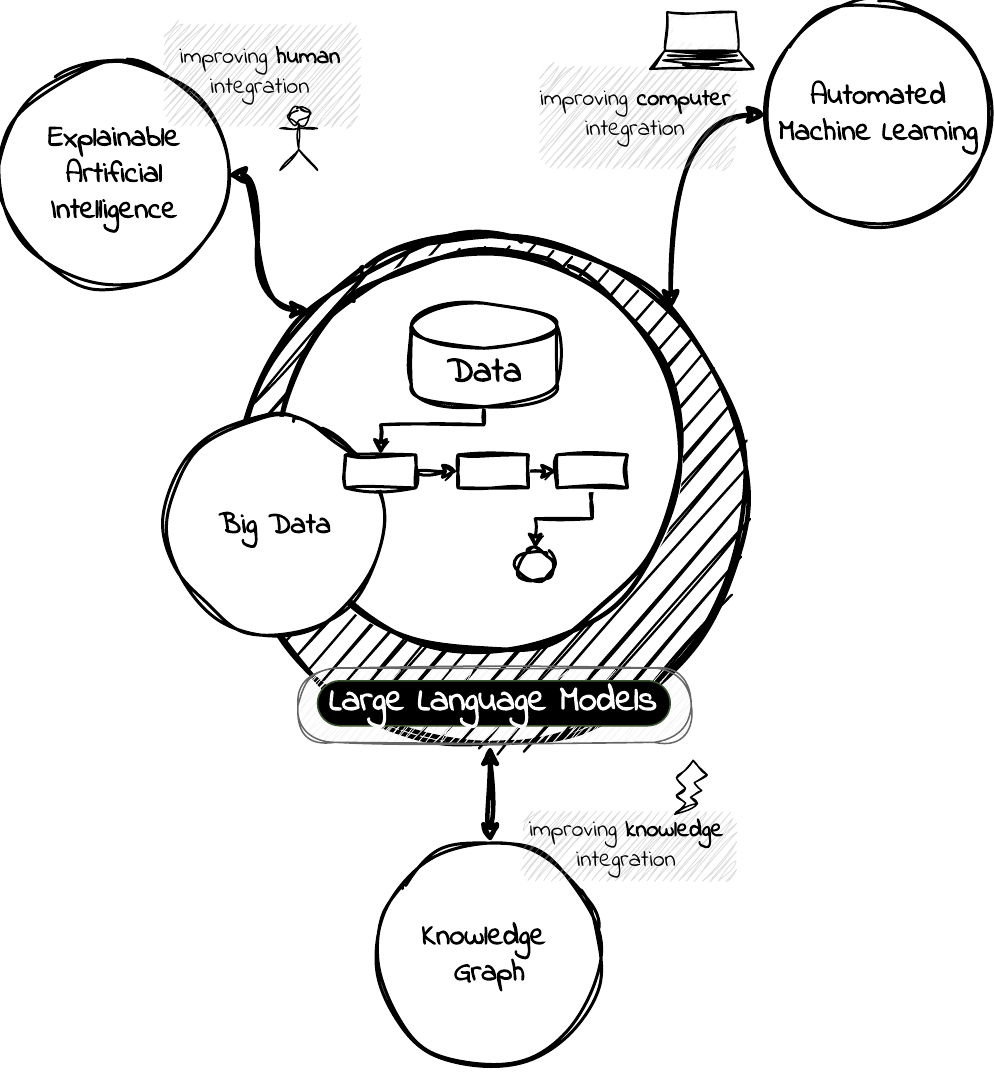}
    \caption{LLM the interface for data pipelines and their cooperation to XAI, AutoML and KG.}
    \label{fig:overview}
\end{figure}

Large Language Models (LLMs) are undeniably pervasive in our lives, serving as copilots, assistants, translators, and reviewers, simplifying tasks from the practitioner's perspective. They have revolutionised the way we access information, offering unparalleled flexibility and democratising access to knowledge.\\
Despite the significant changes and new opportunities offered by LLMs, we believe that their capabilities are far from being fully explored. Furthermore, while LLMs have the potential to serve as interfaces for data pipelines, their adoption and integration into data management systems should be carefully assessed based on specific use cases, requirements, and constraints.\\
Currently, several AI-related fields can cooperatively leverage and advance LLM solutions regarding the mentioned challenges. These fields include eXplainable Artificial Intelligence (XAI), Automated Machine Learning (AutoML), and Knowledge Graphs (KGs) supported by the constraints of Big Data Analytics (BDA).\\
As depicted in Figure~\ref{fig:overview}, we contend that the cooperation and synergy among LLMs, XAI, AutoML, and KGs are crucial for their development and deployment, and are supported by BDA constraints \cite{bellomarini2017swift,ding-etal-2018-improving,chen2020review,gade2020explainable}. 
Big Data supports all mentioned technologies by providing the necessary data resource structures, insights, scalability, and performance required for their development, optimisation, and deployment in various applications and domains. 
\\
From our point of view, XAI can enhance the transparency and understandability of LLMs by providing explanations for their behaviours and outputs. LLMs, in turn, can automate various aspects of XAI tasks by generating natural language descriptions of model predictions or decisions. Integrating both solutions as interfaces to data pipelines can enhance data-related tasks in terms of transparency and flexibility.\\ 
The cooperation between LLMs and AutoML can streamline the data pipeline built upon the machine learning core, automate various tasks, and make machine learning-based systems more accessible to a broader range of users. On the other hand, LLMs can enhance their automation using AutoML systems, optimising pre-training, fine-tuning, and even automating context updating for particular scenarios and users.\\
The synergy between LLMs and KGs empowers data pipeline synthesis by providing a deeper understanding of data, enabling context-aware construction, facilitating intelligent recommendations, and supporting automated optimisation. This integration enhances the efficiency and effectiveness of data pipeline development, ultimately leading to more robust and scalable data processing solutions.\\ 
In the next sections, we provide an overview of LLMs and possible integrations with the aforementioned technologies. Finally, we discuss the main challenges from our point of view towards future directions and conclusions.

\section{Overview of LLMs}
\label{sec:LLMs}
While we lack a precise definition for LLMs~\citep{chang2023survey}, the term typically refers to models that have been extensively pre-trained on large datasets, often consisting of billions of parameters. These models have the remarkable ability to tackle a wide range of linguistic tasks with remarkable skill.\\
Originated from probabilistic language models~\cite{bengio2000neural}, today the term LLMs encompasses a wide variety of models, including but not limited to architectures such as the GPT (Generative Pre-trained Transformer) series, BERT (Bidirectional Encoder Representations from Transformers), and T5 (Text-To-Text Transfer Transformer). What sets these models apart is their ability to learn from massive amounts of text data, capturing the intricate patterns and nuances of language.\\
By pre-training on massive datasets, LLMs acquire a broad understanding of language structure, semantics, and context, enabling them to excel at various natural language processing (NLP) tasks. These tasks range from language translation and sentiment analysis to text summarization and question answering, demonstrating the versatility and power of these models across domains.\\
In addition, the sheer scale of LLMs facilitates their adaptability to different applications and domains. Fine-tuning, a process in which a pre-trained model is further trained on specific data for a particular task, allows for customization according to the requirements of different applications. This flexibility has led to the widespread adoption of LLMs in industries ranging from healthcare and finance to entertainment and education.
LLMs are incredibly potent instruments that leverage transformer architectures with billions or trillions of parameters to capture the intricacies of vast linguistic datasets. By using a two-stage training approach that entails extensive pre-training and targeted fine-tuning, LLMs like OpenAI's GPT-3 have gained recognition for their exceptional generative abilities and adaptability across language tasks \cite{NEURIPS2023_c1f7b1ed}.\\
Recently frameworks like AutoGen~\citep{AutoGen} and LangGraph~\cite{LangGraph} for easy development of LLM applications have been introduced to configure and run multiple agents organised in a \emph{network of LLM agents}, human inputs, and tools conversing with each other. With these frameworks, agent interaction behaviours within LLM applications can be flexibly defined by programming conversation patterns using natural language and computer code \citep{AutoGen}. \\
The authors of \cite{AutoGen} discuss the benefits of their framework for LLM applications like math-problem solving, retrieval-augmented chat, ALF chat, multi-agent coding, dynamic group chat, and conversational chess. In the context of \cite{LangChain} and \cite{LangGraph} there have been several interfaces realised for natural language querying (NLQ) of databases. One of the more sophisticated ones is the Ontotext GraphDB~\cite{NLPOntoText}. For example, it supports queries in natural language text, generates the corresponding SPARQL queries based on a given ontology including trying to self-correct generated erroneous SPARQL queries, and outputs the answer in natural language text. It also supports query and result explanation, summarization, rephrasing, and translation. Interplaying with a vector database, it can index KG entities in a vector database and automatically synchronize changes in RDF data to the KG entity index.\\
These first examples demonstrate the potential of LLM applications and one may get some ideas in which areas and in which way LLM applications can advance the state-of-the-art in other domains as well. In the following section, we introduce our vision of LLMs and LLM applications in data pipelines of KGs, XAI, and AutoML processing Big Data.\\
These adaptable models have a diverse range of applications, including content creation, natural language understanding, sentiment analysis, and multilingual translation. However, it is crucial to address issues such as biases in training data and ethical concerns to ensure the responsible integration of SLMs into the broader landscape of artificial intelligence, as will be discussed in Section \ref{sec:challenges}. Academic discourse is presently exploring subjects such as interpretability, fairness, and ethical governance to refine LLMs for ethical and responsible use. Now, it is imperative to transition these advancements into practical applications within the industry.
\\
\section{Interface for Data Pipelines}
LLMs have the potential to revolutionize data pipelines by providing a more intuitive and user-friendly interface for interacting with and processing data. Their ability to understand and generate natural language makes them valuable tools for democratising access to data and empowering users to derive insights and make informed decisions. Furthermore, the combination of LLMs with XAI, KGs, AutoML, supported by BDA infrastructure enables the development of more powerful and intelligent data pipelines. We discuss this integration for each pair of technologies in the next subsections.
\subsection{Big Data}
Traditional data modelling approaches have become obsolete in the face of the distributed processing demands of Big Data. This paradigm shift requires technologies and strategies that can scale to process and analyse massive data sets. At the same time, this evolution introduces new challenges that require innovative techniques to address issues such as data quality, integration, metadata discovery, and explainable analytics~\citep{ceravolo2018big}.\\
LLMs can make a significant contribution in this area due to their remarkable ability to link data, schemes, and queries to real-world concepts. Unlike traditional models tailored to specific tasks, LLMs offer a broad range of capabilities, including text classification, data generation, and data format translation. This versatility enables them to seamlessly integrate intelligence throughout the data lifecycle, effectively addressing complex database challenges such as entity resolution, schema matching, data discovery, and query synthesis. They also streamline interactions with complicated data structures. Data scientists can express their queries in natural language and translate them into robust search queries across vast repositories~\citep{fernandez2023large}.\\
While the integration of LLMs offers promising benefits, it also raises significant concerns about the reliability of largely automated data processing pipelines~\citep{nascimento2023gpt}. For example, while ChatGPT demonstrates superior problem-solving skills when compared to novice programmers for simple and moderate tasks, it falls short when compared to expert programmers~\citep{nathalia2023artificial}. Moreover, even the most advanced text-to-SQL models, such as GPT-4, achieve only 54.89\% accuracy in execution, a considerable distance from the human benchmark of 92.96\%, proving that challenges still stand~\citep{li2024can}. In \cite{mohta2023large}, it is observed that models trained with human annotations consistently outperform those trained with labels generated by LLMs. Moreover, LLMs exhibit a high demand for computational resources and energy consumption. Such requirements can limit their applicability in certain contexts, necessitating the exploration of optimization strategies to improve operational efficiency~\citep{de2023growing}.\\
To address these challenges, integrating domain-specific knowledge into LLMs through techniques such as LLM fine-tuning and prompt engineering is crucial. This improves the models' ability to handle complex datasets and tasks and reduces biased responses~\citep{zhou2024llm}. Furthermore, \cite{li2024quantity} proposes a self-guided method that uses the representational properties of the target model to identify high-quality data for training. Another way to improve performances is enabling multi-round inference and pipeline execution allows for iterative refinement. Nevertheless, the identification of data quality from LLMs remains a subject of ongoing research. For example, a great interest is arising in developing incremental custom datasets~\citep{10.1007/978-3-031-46846-9_1}.
\subsection{Knowledge Graphs}
As \textit{unifying abstractions}, graphs provide a powerful framework for representing, exploring, predicting, and explaining phenomena in both the real and digital worlds. Leveraging this interconnectedness, graph processing will become not only more pervasive, but also more sophisticated, with diverse frameworks, algorithms, and infrastructure designed to support the analysis and manipulation of big graphs~\citep{sakr2021future,bellandi2022toward}. \\
KGs, as structured representations of knowledge consisting of entities, attributes, and relationships, play a critical role in graph processing ecosystems by organizing and linking diverse information sets in a machine-readable format~\citep{hogan2021knowledge}. LLMs will further enhance the capabilities of graph processing ecosystems by facilitating a seamless interface between complex data structures and natural language interfaces~\citep{Pan_2024}.\\
The synergy between LLMs and KGs holds great promise as combining them creates more informative and comprehensive AI systems. \cite{Pan_2024} provides a roadmap for integrating LLMs and KGs, including KG-enhanced LLMs, LLM-augmented KGs and synergized LLMs and KGs. 
On the one hand, KGs can act as factual knowledge reducing the occurrence of misinformation and hallucinations, to boost the LLM precision; and can be incorporated in various stages: (i) the LLM pre-training to improve knowledge representation; (ii) the LLM inference to enable access to current knowledge without retraining; and the LLM’s interpretability to understand LLM reasoning. 
On the other hand, LLMs can aid in extracting information (such as named entities and relationships) from text to populate knowledge graphs and enhance other KG tasks including embedding and completion. \cite{ding-etal-2018-improving} demonstrate the effectiveness of simple constraints in improving KG embedding, while \cite{chen2020review} highlight the potential of LLMs in KG construction and reasoning. For KG completion, KG-BERT \cite{yao2019kg} is a leading example of applying LLMs to identify missing triples in KGs. There are also LLM-based methods to detect inconsistencies in KGs, such as AutoAlign \cite{zhang2023autoalign} that leverages LLMs to automatically and accurately aligning entity types of two KGs. 
LLMs can also act as a user-friendly knowledge access for querying knowledge graphs, allowing users to ask questions and receive summaries in natural language without the need for complex query languages. 
Furthermore, \cite{bellomarini2017swift} emphasises the importance of efficient and scalable reasoning over big data in the context of KG management systems, which underscores the foundational role of big data constraints in enabling the cooperation and synergy among LLMs, XAI, AutoML, and KGs.

\subsection{Explainable Artificial Intelligence}
Explainability refers to the ability to provide transparent and interpretable insights into an algorithm's decision-making process, enabling stakeholders to understand the factors, features, and logic that influence its outputs~\citep{nauta2023anecdotal}. It involves revealing the internal workings, data dependencies, and decision paths of the algorithm to foster trust, accountability, and understanding among users and stakeholders~\citep{arrighi2023explainable}. Interpretability and intelligibility are alternative terms used in the literature to refer to these concepts.\\
LLMs hold great promise for advancing explainability efforts. Using their natural language processing skills, it is possible to elicit clear and contextually appropriate explanations for the results generated by complex algorithms and predictive models. In addition, LLMs excel at contextualizing these explanations to enrich the information conveyed. They also excel at synthesizing complex data and distilling it into key insights. In addition, LLMs can provide illustrative examples and counter-examples to demonstrate the impact of changes in input data on model predictions, giving users a clearer understanding of how the model will respond in different scenarios.\\
Recent research has explored the use of LLMs in XAI to improve model explanations. \cite{mavrepis2024xai} developed a custom GPT model, x-[plAIn], that can generate concise and easily understandable summaries of complex XAI methods, tailored to suit the varying expertise levels and interests of different audience groups. LLMs can be prompted or fine-tuned to provide answers using a ``Chain of Thought''~\citep{wei2022chain}. Unlike traditional methods where the model directly outputs the final answer, the Chain of Thought approach encourages the model to generate intermediate reasoning steps that lead to the final answer.\\
\cite{chen2023lmexplainer} proposed LMExplainer, a knowledge-enhanced explainer for LLMs that outperformed existing methods in providing human-understandable explanations. When presented with a question context \textit{z}, comprising the question \textit{q} and a set of answers \textit{A}, the proposed approach begins by generating language embeddings using a pre-trained LLM. Concurrently, it retrieves relevant knowledge from a KG to construct a subgraph. The language embeddings and subgraph are then merged to derive Graph Neural Network (GNN) embeddings. These combined embeddings are subsequently fed into a Graph Attention Network (GAT) to compute attention scores. These attention scores serve a dual purpose. Firstly, they assess the significance of the GNN embeddings, which are then utilised alongside the language embeddings for the final prediction. Secondly, they facilitate the generation of explanations by highlighting the most critical aspects of the reasoning process.\\ 
Taken together, these studies underscore the potential of LLMs to improve the accessibility and power of XAI. The hope is that as LLMs continue to evolve and improve, they are likely to play a key role in democratizing XAI, making it more accessible and useful in different domains and applications. However, the integration of LLMs into XAI also poses significant challenges. These challenges include the reproducibility of results, ensuring the robustness and fairness of the generated explanations, and mitigating biases inherent in the training data and model outputs. \\
\subsection{Automated Machine Learning}
AutoML refers to the process of automating the end-to-end process of applying machine learning techniques to real-world problems. It aims to make machine learning more accessible to non-experts by automating various tasks involved in the machine learning pipeline, such as data preprocessing, feature selection, model selection, hyperparameter tuning, and model evaluation \citep{hutter2019automated}.
Integrating LLMs into AutoML platforms can enhance the automation of machine learning tasks. LLMs can assist in automating various aspects of the most traditional tasks of AutoML such as Algorithm Selection (AS), Hyperparameter Optimisation (HPO), Combined Algorithm Selection and Hyperparameter Optimisation (CASH), Neural Architecture Search (NAS)  and Pipeline Synthesis (PS) ~\citep{hutter2019automated,he2021automl,zheng2023automl}. On the other hand, as mentioned by \cite{tornede2024automl}, AutoML for LLMs presents unique challenges compared to previous applications of AutoML to different learning paradigms. Standard AutoML solutions are deemed partially ineffective for LLMs due to their resource-intensive nature, as argued by \cite{godbole2023deep}. Moreover, the demand for innovative solutions to automatically obtain LLMs, with comparable or superior quality to manually designed models, need to be addressed in the next generation of deep learning methods.\\
Considering how LLMs can boost AutoML task, we consider that AS and CASH can take advantage of LLMs by analysing the characteristics of a given input dataset and the specific requirements of the task described in natural language. They can understand user queries or descriptions of the problem and recommend suitable algorithms based on their understanding of different machine-learning techniques and their performance on similar tasks.\\
LLMs can help HPO by generating natural language descriptions of the desired model performance and constraints. LLMs can also describe the impact of different hyperparameters on model performance, helping users to comprehend the decisions during the optimisation process. Similarly, for PS tasks, LLMs can assist by generating natural language descriptions of the desired data processing steps and model architectures.\\
Investigating the potential of applying AutoML techniques to LLMs, particularly NAS, for optimising pre-training and fine-tuning presents exciting opportunities for advancing the capabilities of LLMs. The existing literature~\citep{tornede2024automl,godbole2023deep} highlights five significant challenges: a) Pre-training LLM base models are excessively costly, restricting the number of complete training iterations; b) AutoML tasks for LLMs are intricate and encompass various stages of the lifecycle, yet current methodologies lack comprehensive optimization across these stages; c) Discovering optimal neural architectures for LLMs remains difficult, despite automated methods like NAS, which have shown limited advancements in this area; d) Each phase of the LLM lifecycle demands optimization of distinct metrics, potentially resulting in misleading performance indicators for AutoML; e) AutoML traditionally concentrates on a single learning paradigm while training LLMs involves navigating multiple paradigms throughout the lifecycle.

\section{Challenges and Considerations}
\label{sec:challenges}

Can we envision a future where data scientists and experts in machine learning or artificial intelligence are no longer needed, or will they remain essential only for the development of LLMs, AutoML, XAI systems? \\
The answer is quite speculative and complex. However, it's important to consider factors such as advancements in automation (AutoML), the complexity of problems, ethical and social implications, continuous innovation and adaptation, human judgment and Interpretability. Considering these factors, including the LLMs improved interface, while automation may streamline certain aspects of AI development and deployment, it's unlikely that data scientists and AI experts will become obsolete in the foreseeable future. Instead, their roles may evolve, focusing more on high-level problem-solving, ethical considerations, interdisciplinary collaboration, and innovation, while leveraging automation tools to enhance productivity and efficiency. Ethical considerations play an important role in the integration of new technologies (e.g., LLMs). These models have the potential to perpetuate biases present in the training data or make decisions with far-reaching consequences. Care must be taken to mitigate these risks and ensure fair and equitable outcomes.
\\
While the ease of using technologies for certain tasks has never been seen before with LLMs, it also raises concerns about energy consumption. For instance, while a pocket calculator can run for years on an LR44 battery offering 0.0002325 KWh, just one average Google search consumes even more energy\footnote{\url{https://techland.time.com/2011/09/09/6-things-youd-never-guess-about-googles-energy-use/} (visited on 15.3.2024)}: 0.0003 KWh. Depending on the size of the model and the number of tokens processed, some estimates\footnote{\url{https://lifestyle.livemint.com/news/big-story/ai-carbon-footprint-openai-chatgpt-water-google} \url{-microsoft-111697802189371.html} (visited on 15.3.2024)} suggest that the energy consumption of a ChatGPT-4 query ranges from 0.001 to 0.01 KWh, which is 3.33 to 33.33 times more energy than a Google search, equivalent to the energy of 4 to 43 LR44 batteries. As the revolution in programming and automated problem-solving occurs daily, climate change concerns demand a reevaluation of the unreflected use of LLMs for daily tasks and more complex endeavors, as proposed here for AutoML and XAI. We may introduce conventions to use LLMs just once for generating configurations of data pipelines and efficient code to run instead of solving problems instructed in natural language, as proposed in LLM applications (see Section~\ref{sec:LLMs}). Future technologies might automatically detect such prompts and choose the best option to reduce the carbon footprint.
\\
Addressing the challenges associated with emerging technologies like LLMs is paramount. These challenges encompass several key areas, including the prohibitively high cost of pre-training LLM base models, the intricate nature of AutoML and XAI tasks spanning multiple stages of the lifecycle, the complexity in identifying optimal neural architectures, the necessity for optimising diverse metrics throughout the LLM lifecycle, and the importance of navigating multiple learning paradigms while maintaining proper structures such as KG.\\
While the integration of LLMs into new platforms presents promising opportunities for task automation, human-like problem-solving, and enhancing human performance, it's imperative to address the hurdles posed by these new technologies to fully leverage their potential in advancing the fields of NLP and AI.\\
One fundamental similarity among all new technologies, including LLMs, is their inherently open-ended nature, where definitive ground truths are often elusive. Consequently, frameworks for human-centered qualitative evaluation become imperative. While XAI benefits from a well-established field of study focusing on various cognitive aspects, LLMs and AutoML currently lack exploration in this domain, particularly concerning qualitative evaluation methodologies.\\
Enhancing our understanding and addressing these challenges will not only facilitate the responsible development and deployment of LLMs but also contribute to the broader advancement of AI and NLP. It underscores the importance of interdisciplinary collaboration, ethical considerations, and ongoing innovation to harness the transformative potential of these technologies while ensuring their alignment with societal values and needs.\\
\section{Future Directions and Conclusions}
LLMs offer unprecedented simplicity for certain tasks, raising questions about the future role of data scientists and data engineers. While they have the potential to revolutionise AI applications, they also present challenges in areas ranging from ethics to energy consumption. Future technologies, mainly XAI, AutoML, and KG, supported by BDA constraints, may need to establish conventions and interactions to optimise LLM usage and ensure their optimal performance.\\
Despite the immense potential of LLMs, responsible integration and addressing associated challenges are essential to harness their benefits effectively. Ethical considerations must remain central in AI development and deployment, guiding decisions about LLM usage and mitigating risks such as biases, errors, and misinformation that may arise from inadequate quality control of training datasets.\\
By prioritising ethical considerations in the integration and application of LLMs with other AI technologies, we can maximise their positive impact while mitigating potential risks, thus paving the way for a more ethical and inclusive AI-driven future.

\bibliographystyle{ACM-Reference-Format}
\bibliography{sigproc.bib}

\end{document}